\documentclass{article}

\usepackage[nonatbib, final]{nips_2016}

\usepackage[utf8]{inputenc} % allow utf-8 input
\usepackage[T1]{fontenc}    % use 8-bit T1 fonts
\usepackage{hyperref}       % hyperlinks
\usepackage{url}            % simple URL typesetting
\usepackage{booktabs}       % professional-quality tables
\usepackage{amsfonts}       % blackboard math symbols
\usepackage{nicefrac}       % compact symbols for 1/2, etc.
\usepackage{microtype}      % microtypography

\usepackage[english]{babel}
\usepackage{amsmath}
\usepackage{array}

\usepackage{graphicx} % more modern
\usepackage[numbers]{natbib}
\newcommand{\bs}[1]{{\boldsymbol{#1}}}

\newcommand{\db}[1]{{[\![ {#1} ]\!]}}

\title{Predictive Clinical Decision Support System with RNN Encoding and Tensor Decoding}

\author{
  Yinchong Yang \\
  Ludwig-Maximilians-Universität München, Munich,  Siemens AG, Corporate Technology, Munich \\
  \texttt{yinchong.yang@siemens.com} 
  \And
  Peter A. Fasching \\
  Department of Gynecology and Obstetrics, University Hospital Erlangen, \\Comprehensive Cancer Center Erlangen-EMN, \\Friedrich-Alexander University Erlangen-Nuremberg, Erlangen, Germany  \\
  \texttt{peter.fasching@uk-erlangen.de} \\
  \And
   Markus Wallwiener \\ 
   Department of Gynecology and Obstetrics, University Hospital Heidelberg, Heidelberg, Germany \\
  \texttt{markus.wallwiener@med.uni-heidelberg.de}
  \And
  Tanja N. Fehm \\ 
  Department of Gynecology and Obstetrics, University Hospital Düsseldorf, \\Heinrich-Heine University Düsseldorf, Düsseldorf, Germany \\
  \texttt{tanja.fehm@med.uni-duesseldorf.de}
  \And
  Sara Y. Brucker \\ 
  Department of Gynecology and Obstetrics, University Hospital Tübingen, Tübingen, Germany \\
  \texttt{Sara.Brucker@med.uni-tuebingen.de}
  \And
  Volker Tresp \\
  Ludwig-Maximilians-Universität München, Munich, Siemens AG, Corporate Technology, Munich \\
  \texttt{volker.tresp@siemens.com} \\
}

\begin{document}

\maketitle

\section{Introduction}

With the introduction of the Electric Health Records (EHR), large amounts of digital data become available for analysis and decision support \cite{rahman2015electronic}. 
These data, as soon as carefully cleaned and well preprocessed, could enable a large variety of analysis and modeling tasks that can improve healthcare services and patients experience \cite{tresp2016going}. 
When physicians are prescribing treatments to a patient, they need to consider a large range of data variety and volume. 
These data might include patients' genetic profiles and their entire historical clinical protocols. With the growing amounts of data decision making becomes increasingly complex.
Machine learning based Clinical Decision Support (CDS) systems can be a solution to the data challenges~\cite{choi2015doctor} \cite{esteban2015predicting} \cite{esteban2016predicting}.
Machine learning models and decision support systems have been proven to be capable of handling ---and actually even profiting from--- large amount of data in high dimensional space and with complex dependency characteristics.
 Some powerful machine learning models generate abstract and yet informative features from a usually sparse feature space.

There are multiple ways that a machine learning model may impact the decision process of a physician: either indirectly, by predicting the possible outcome of each decision; or directly by calculating recommendation scores for all possible actions. 
As an example of the former case, \cite{esteban2016predicting} provides physicians with endpoint predictions of each patient queuing for a kidney transplantation. Based on the predicted probabilities of kidney rejection, kidney loss and death, within the next 6 and 12 months, the physician may decide which candidate should receive a donated kidney. 

\begin{figure}[h]
	\centering
	\caption{The concept of a machine learning based CDS system. The system, as long as sufficiently trained on historic data, could predict the physician's decision set. }
	\label{fig:concept_DCS}
	\includegraphics[clip, trim=0cm 1cm 0cm -1cm, scale=0.33]{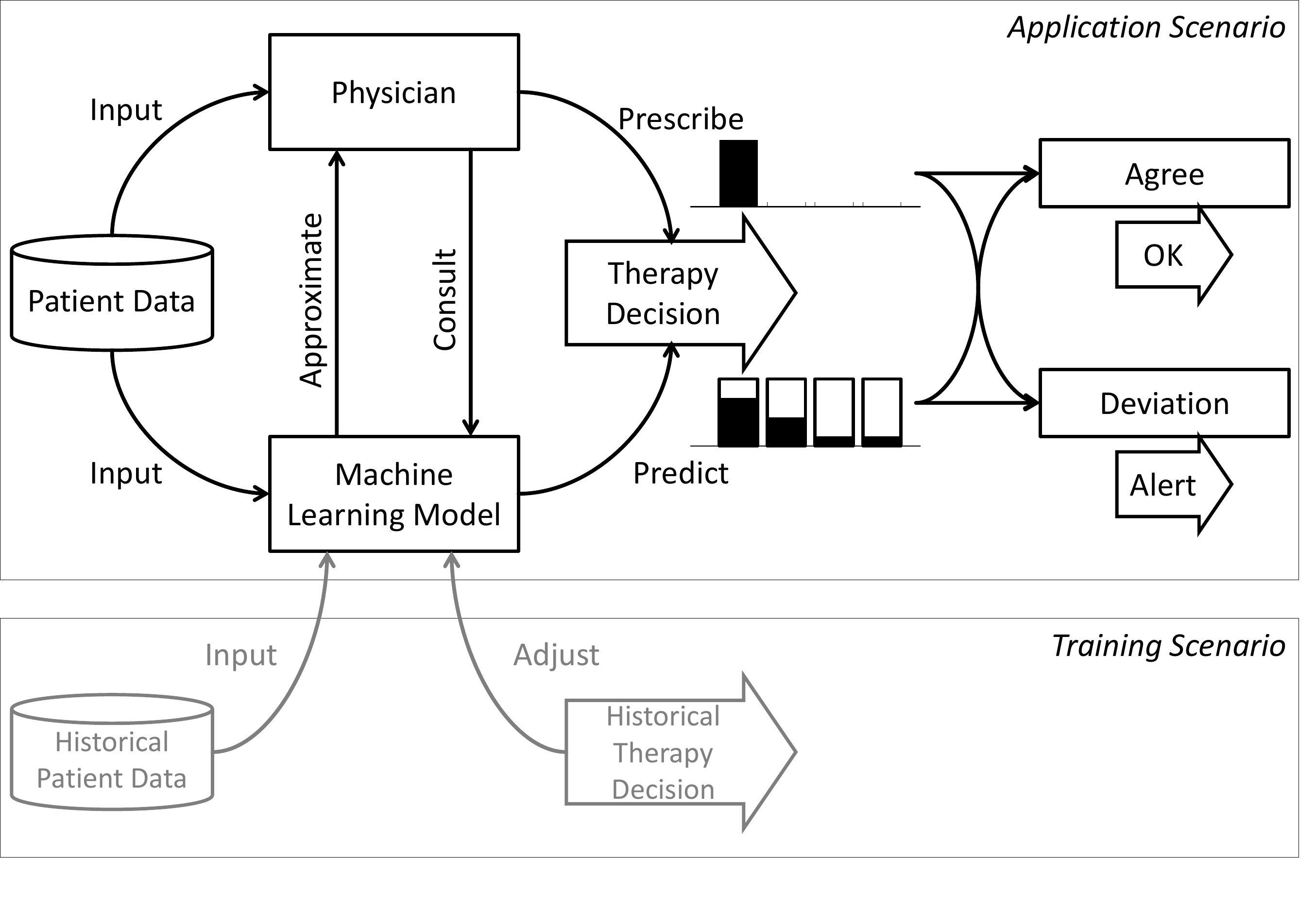}
\end{figure}

In this work we focus on another other class of decision support in which the physicians' decision is directly predicted. Concretely, the model would assign higher probabilities to decisions that it presumes the physician are more likely to make. Thus the CDS system can provide physicians with rational recommendations. As a concrete use case, as soon as a physician prescribes a treatment that is not in the top-$n$ ranked recommendations made by the CDS system, an alert would be triggered to ask the physician to reconsider the prescription and/or to document the arguments for the decision in more details. This scenario is illustrated in the upper part of the Figure \ref{fig:concept_DCS}.

The proposed system is based on the predictive power of machine learning models, which are trained using historical data as illustrated in the lower half of the Figure \ref{fig:concept_DCS}. During training, the model attempts to predict historical decisions based on corresponding patient data. The actually documented decisions can adjust the model so that it can improve its predictions throughout the training epochs.

In our work we also address a problem that has not yet drawn much attention in the designing of a CDS, i.e. that  a physician is required to make multiple decisions in a block, say, the intention and the type of a radiotherapy, and that these decisions are mutually dependent. In machine learning, this is known as the issue of correlation in target features. We propose a solution to the target correlation problem using a tensor factorization model.

In order to handle the patients' historical information as sequential data, we apply the so-called Encoder-Decoder-Framework which is based on Recurrent Neural Networks (RNN) as encoders and a  tensor factorization model as a decoder, a combination which is novel in machine learning.

% Volker-Comment: sometimes these correlated decisions are called an ``order set'' in healthcare

\section{Decision Support for Breast Cancer Therapies} \label{sec:main_section}

\subsection{Data Description}

The data, provided by the PRAEGNANT study network\cite{fasching2015biomarker}, were collected on patients suffering from metastatic breast cancer. 

After preprocessing we could extract information on $1245$ patients: The 1) basic information, 2) primary tumor and 3) history of metastasis before entering the study. These data are considered to be static and consist in total of 26 features of binary, categorical or real types. We performed dummy-coding on the former both cases and could extract for each patient $i$ a static feature vector denoted with $\bs{m}_i \in \mathbb{R}^{114}$

The sequential information consists of two categories. Firstly, 4) local recurrences, 5) metastasis and 6) visits cover information on the clinical events, such as progression and live status, to which we refer as \emph{events}. Secondly, there are three types of \emph{therapy} informations: 7) radiotherapies, 8) systemic therapies and 9) operations. These sequential data, being only either binary or categorical, are also dummy-coded. The total number of actual events for each patient varies from $1$ to $23$ and is on average $5$. We retrieve for each patient $i$ at a time step $t$ a feature vector $\bs{x}_{i}^{[t]} \in \{ 0, 1 \}^{182}$ and denote the whole sequence for patient $i$ up to time $t$ using a set of $\{ \bs{x}_{i}^{[t']}\}_{t'=1}^{t}$.
% Yinchong-Comment: "sequential information" consists first of two groups: events and therapies. The first consists of 4), 5) and 6); while the latter consists of 7), 8) and 9).

Based on the static and all the sequential information up to a time step $t-1$, we attempt to model how the radiotherapy at $t$ should be prescribed, which consists of two 3-D feature vectors: The first one is the therapy intention, being either curative, palliative or unknown. The second target feature is the therapy type, being either percutaneous, Brachytherapy or unknown. We denote these as $\bs{y}_{\langle i,t \rangle} \in \{0,1\}^3$ and $\bs{z}_{\langle i,t \rangle} \in \{0,1\}^3$ for patient $i$ at time $t$. We perform Pearson's $\chi$-squared tests and G-Tests on the two target features and could verify the existence of correlations, with test-statistics of $\chi^2=197.17$ and $G=146.48$ respectively (DF=4). The $p$-values in both cases are less than 2.2e-16 according to the implementation in the statistical programming framework of R \cite{rcore} \cite{desctools}.
% Yinchong-Comment: the target has two 3-D feature vectors, one describing the radiotherapy's type and another its intention.

\subsection{Modeling Solution} \label{subsec:modeling_solution}

First we associate the sequential and static data on a patient $i$ up to a time step $t$ by constructing a latent representation vector:
\begin{align}
	\bs{a}_{\langle i,t \rangle} = ( g(\{ \bs{x}_{i}^{[t']}\}_{t'=1}^{t}) ~|~ \sigma( \bs{H}^{T} \bs{m}_i ) ),
\end{align}
where $g$ denotes an RNN encoder that outputs only the last hidden state given the input sequence $\{ \bs{x}_{i}^{[t']} \}_{t'=1}^{t}$. This approach was first proposed by \cite{sutskever2014sequence} where such RNN is proven to be able to encode all information in a sequence of variable length into a single fixed size real vector. We exploit this aspect of such RNN encoders and apply this vector to represent the patient-specific temporal information indexed by $\langle i, t \rangle$. We augment this information by concatenating it with with the static patient information, which is log-linearly encoded by a matrix $\bs{H}$ and the sigmoid activation function~\cite{esteban2016predicting}.

In order to take into account the verified correlation between target features we propose to model the probability of each possible \emph{pair} of target feature values, corresponding to the \emph{joint} probability distribution of each feature value. The probability of observing a value pair $\langle$intention$=j$ ,type$=k$$\rangle$ given the static and sequential input data would be calculated as:
\begin{align}
	\mathbb{P}( ({\bs{y}}_{\langle i, t \rangle})_j &= 1 \wedge ({\bs{z}}_{\langle i, t \rangle})_k = 1 | \bs{m}_i, \{ \bs{x}_{i}^{[t']}\}_{t'=1}^{t} ) =: ( \bs{U}_{\langle i, t \rangle} )_{j, k} ~~\text{where}  \label{eq:joint_prob} \\
	\bs{U}_{\langle i, t \rangle} &= \bs{y}_{\langle i, t \rangle} \otimes \bs{z}_{\langle i, t \rangle} \in \{0, 1\}^{3 \times 3}, \label{eq:outer_prod}
\end{align}
where we construct a new target feature \emph{matrix} $\bs{U}$ by calculating the outer product of the dummy coded feature \emph{vectors}. Therefore each entry in the matrix represents the probability that a certain pair of target feature values is observed on these two features. Considering also the $\langle$patient, time$\rangle$-dimension we have a 3-way tensor as modeling targets.

Now we construct our regression model toward the target tensor in a fashion similar to Tucker factorization:
\begin{align} \label{eq:ree_model}
		\hat{u}_{\langle i, t \rangle, j, k} &= \sigma( \db{\bs{\mathcal{G}}; ~\bs{a}_{\langle i,t \rangle}, \bs{b}_j, \bs{c}_k  } = \sigma( \bs{a}_{\langle i,t \rangle}^{T} \bs{G}_{(1)} vec( (\bs{c}_k \otimes \bs{b}_j)^{T} ) )
\end{align}
where $\bs{G}_{(1)}$ represents the unfolding of tensor $\bs{\mathcal{G}}$ w.r.t the $1$st dimension following \cite{kolda_tensor_2009}.

\begin{figure*}[t]
	\centering
	\caption{An overview of the entire model architecture. The latent embedding vector $\bs{a}_{\langle i, t \rangle}$ is derived from the static and sequential information, and is the input to the target tensor.}
	\label{fig:SimplifiedModel}
	\includegraphics[clip, trim=5cm 15.5cm 5cm 20cm, scale=0.45]{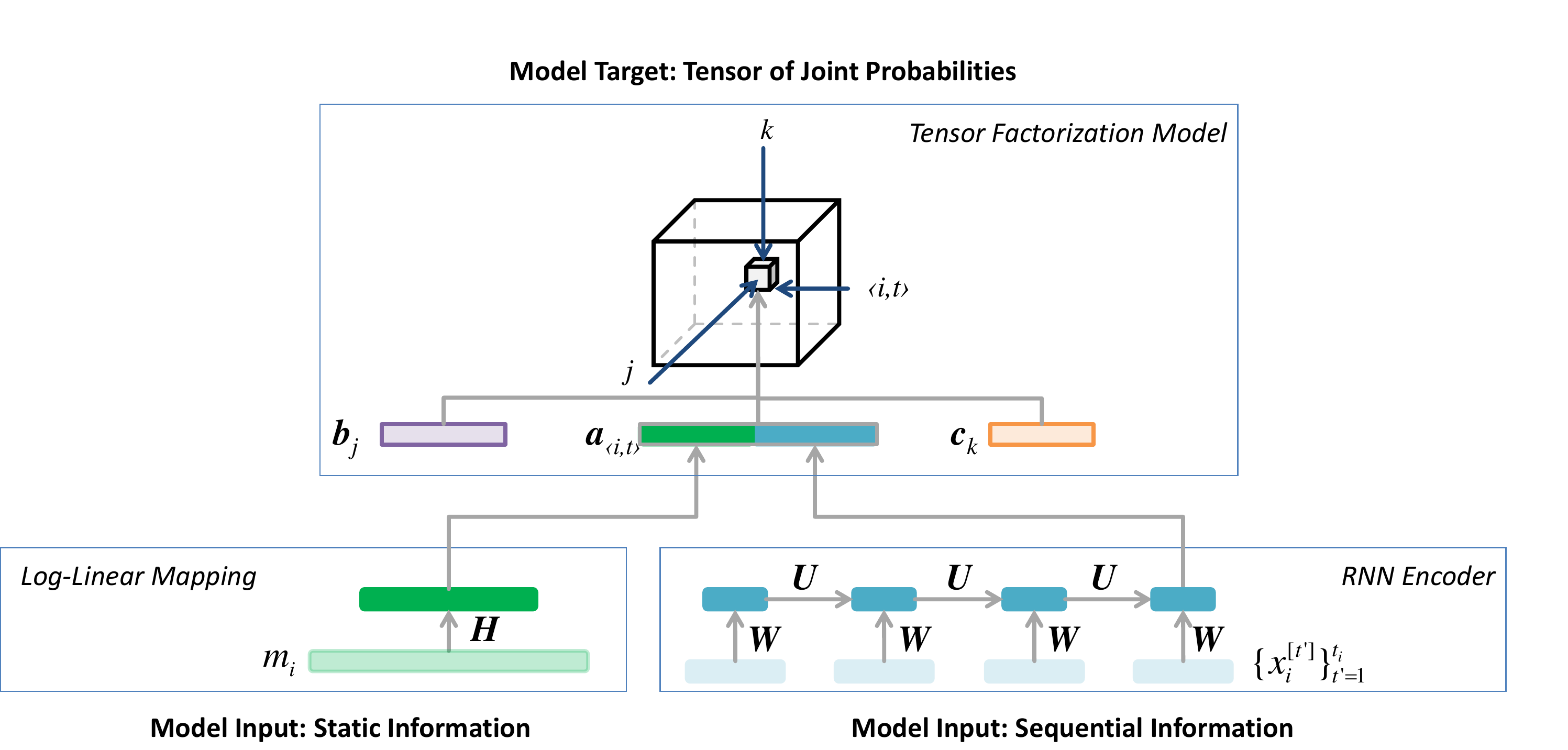}
\end{figure*}

The entire model architecture, illustrated in Figure \ref{fig:SimplifiedModel}, is trained end-to-end.

\section{Experiments}

We assign $80\%$ of all the $1245$ patients to the training set and $20\%$ to the test set. Then we query all the static and sequential information as well as targets belonging to the training patients and test patients respectively. As the RNN model we choose the LSTM with hidden state of size $25$ and map the static information into a latent space of size $15$ using a simple log-linear mapping matrix $\bs{H}$ as in Figure \ref{fig:SimplifiedModel}. As tensor model we implement a rank 5 Tucker-3 model. We apply $0.1$-dropout regularization for the RNN part and $0.01$-ridge regularization for the rest of the parameters in the model. All models are implemented using Theano \cite{Bastien-Theano-2012} \cite{bergstra+al:2010-scipy} and Keras \cite{keras2015}. All models, including the trainable baseline model for the marginal distribution of feature values, are trained up to $1000$ epochs using RMSprop algorithm with learning rate $0.001$.
We sample $20\%$ within the training set to be a validation set for the early-stopping mechanism and generate 5 different splittings of the data to perform cross-validations and report the mean and standard deviation of all 5 predictions for each target group. As objective function we define the sum of all binary cross entropies for each target feature as
\begin{align}
	- \sum _{\forall i} \sum_{t=1}^{T_i} &\left[ \sum_{j=1}^{3} \sum_{k=1}^{3} \delta \left( u_{\langle i, t \rangle, j,k}, ~\hat{u}_{\langle i, t \rangle,j,k} \right)  \right] ~\text{with}~ \delta(y, \hat{y}) = y \cdot \log(\hat{y}) + (1-y) \cdot \log(1 - \hat{y})
\end{align}

As baseline models we test 1) random guessing; 2) most popular prediction, which is to constantly predict the frequency of each target feature value in the training data; and 3) a standard model that predicts the marginal distributions. Since our tensor model predicts the joint distribution, the predictions of both models are not directly comparable. We calculate the outer product of the marginal distributions from the baseline models, which would serve as a pseudo joint probability with which we can compare against our tensor model.

We evaluate the performances of all models in term of the AUROC which is standard for classification tasks. This metric can be here interpreted as the capability of the model to assign a patient group to a specific feature value. In our application scenario, however, we find it of more relevance that the  CDS system can generate for each patient a set of rational therapy recommendations. In order to also test this aspect we evaluate the models using ranking-based metrics such as Coverage Error \cite{tsoumakas2009mining} and Ranking Precision \cite{madjarov2012extensive} from the scikit-learn tool-box \cite{scikit-learn} and NDCG@k \cite{shani2011evaluating}.

We report in Table \ref{tab:results} our experimental results.
\begin{table}[h]
\caption{Experimental Results}
\label{tab:results}
	\begin{tabular}{| l | l | l | l | l | l | l |} \hline
		Metrics & Random & Most Popular & Standard Model & Tensor Model \\ \hline \hline
		AUROC & 0.489$\pm$0.014 & 0.587$\pm$0.165 & 0.842$\pm$0.009 & \textbf{0.874}$\pm$0.012 \\ \hline
		Coverage Error & 138.846$\pm$0.740 & 118.569$\pm$11.053 & 28.197$\pm$1.344 & \textbf{27.902}$\pm$1.084 \\ \hline
		Rank Precision & 0.098$\pm$0.008 & 0.079$\pm$0.022 & 0.226$\pm$0.008 & \textbf{0.296}$\pm$0.009 \\ \hline
		NDCG@5 & 0.066$\pm$0.002 & 0.047$\pm$0.031 & 0.172$\pm$0.002 & \textbf{0.269}$\pm$0.014 \\ \hline
	\end{tabular}
\end{table}
We can verify that, with the present experimental setting, the tensor-target model indeed provides better performances by considering the correlations between target feature values. Especially in term of NDCG@5, which is a common metric in assessing recommender systems, the tensor model improves the prediction by as much as $56\%$ .

\section{Discussions}
The major contribution of our work is to propose a novel model architecture by combining an RNN encoder with a tensor factorization model, both of which share a common latent representation. The tensor factorization component models the joint probability of target feature values instead of the marginal ones. We hypothesize that our model is therefore capable of capturing correlations among target features, which is often relevant in modeling clinical decisions. We show that the proposed model does achieve better prediction performances in experiments with real-world datasets. Within this paper we only conduct prediction of the radiotherapy. We shall extend our work to the other two therapies in our dataset: the systemic therapy and operations. Especially the former one has three targets and would require a 4-D tensor.

%\section*{References}
\small
\bibliographystyle{IEEEtran}
\bibliography{references}

\end{document}